\documentclass[10pt,twocolumn,letterpaper]{article}

\usepackage[pagenumbers]{wacv} 

\definecolor{wacvblue}{rgb}{0.21,0.49,0.74}
\usepackage[pagebackref,breaklinks,colorlinks,allcolors=wacvblue]{hyperref}
\usepackage[numbers,sort&compress]{natbib}

\def\wacvPaperID{1151} 
\def\confName{WACV}
\def\confYear{2026}

\newcommand{\myparagraph}[1]{\smallskip\noindent\textbf{#1}}
\newcommand{\myfirstpara}[1]{\noindent\textbf{#1}}
\usepackage{booktabs}

\newcommand{\tenk}{\textsc{$100{,}000 \times 100{,}000$}\xspace}

\usepackage{soul}

\usepackage{multirow}
\usepackage{amssymb}
\usepackage{wrapfig,booktabs}
\usepackage{graphicx}
\usepackage{wrapfig}
\usepackage{threeparttable}

\newcommand{\ua}[1]{$\uparrow#1 $}

\title{Efficient Whole Slide Pathology VQA via Token Compression}

\author{Weimin Lyu\textsuperscript{\textnormal{1}\thanks{Equal contribution.}}, 
Qingqiao Hu\textsuperscript{\textnormal{1}\footnotemark[1]}, 
Kehan Qi\textsuperscript{\textnormal{1}}, 
Zhan Shi\textsuperscript{\textnormal{1}}, 
Wentao Huang\textsuperscript{\textnormal{1}}, 
Saumya Gupta\textsuperscript{\textnormal{1}}, 
Chao Chen\textsuperscript{\textnormal{1}}  \\ 
\textsuperscript{1} Stony Brook University}

\begin{document}
\maketitle
\begin{abstract}

Whole-slide images (WSIs) in pathology can reach up to \tenk pixels, posing significant challenges for multimodal large language model (MLLM) due to long context length and high computational demands. Previous methods typically focus on patch-level analysis or slide-level classification using CLIP-based models with multi-instance learning, but they lack the generative capabilities needed for visual question answering (VQA). More recent MLLM-based approaches address VQA by feeding thousands of patch tokens directly into the language model, which leads to excessive resource consumption. To address these limitations, we propose Token Compression Pathology LLaVA (TCP-LLaVA), the first MLLM architecture to perform WSI VQA via token compression. TCP-LLaVA introduces a set of trainable compression tokens that aggregate visual and textual information through a modality compression module, inspired by the [CLS] token mechanism in BERT. Only the compressed tokens are forwarded to the LLM for answer generation, significantly reducing input length and computational cost. Experiments on ten TCGA tumor subtypes show that TCP-LLaVA outperforms existing MLLM baselines in VQA accuracy while reducing training resource consumption by a substantial margin.

\end{abstract}

\section{Introduction}

Understanding whole slide images (WSIs) at the gigapixel scale has emerged as an increasingly important topic in computational pathology. However, WSIs are typically extremely large, often spanning dimensions of up to \tenk pixels, which poses significant computational challenges for multimodal large language modeling.

Most existing computational pathology methods are limited to patch-level analysis or slide-level classification, restricting their ability to reason effectively over entire WSIs. For example, models like QUILT-LLaVA \cite{seyfioglu2024quilt} and LLaVA‑Med \cite{li2023llava} operate on isolated patches, resulting in fragmented insights that miss the broader context of the entire slide. 
On the other hand, classification-oriented methods typically leverage CLIP backbones~\cite{radford2021learning} combined with multi-instance learning (MIL) to aggregate patch embeddings into slide-level representations. MIL frameworks mitigate diagnostic sparsity by pooling thousands of patch features into a compact representation at the slide level. While models like PathGen‑CLIP~\cite{sun2025pathgenm} achieve strong classification performance through contrastive training on image-text pairs, their outputs remain limited to classification tasks and lack generative capabilities required for text-based tasks such as visual question answering (VQA).





\begin{figure*}[!t]
  \centering
  \includegraphics[width=0.8\textwidth]{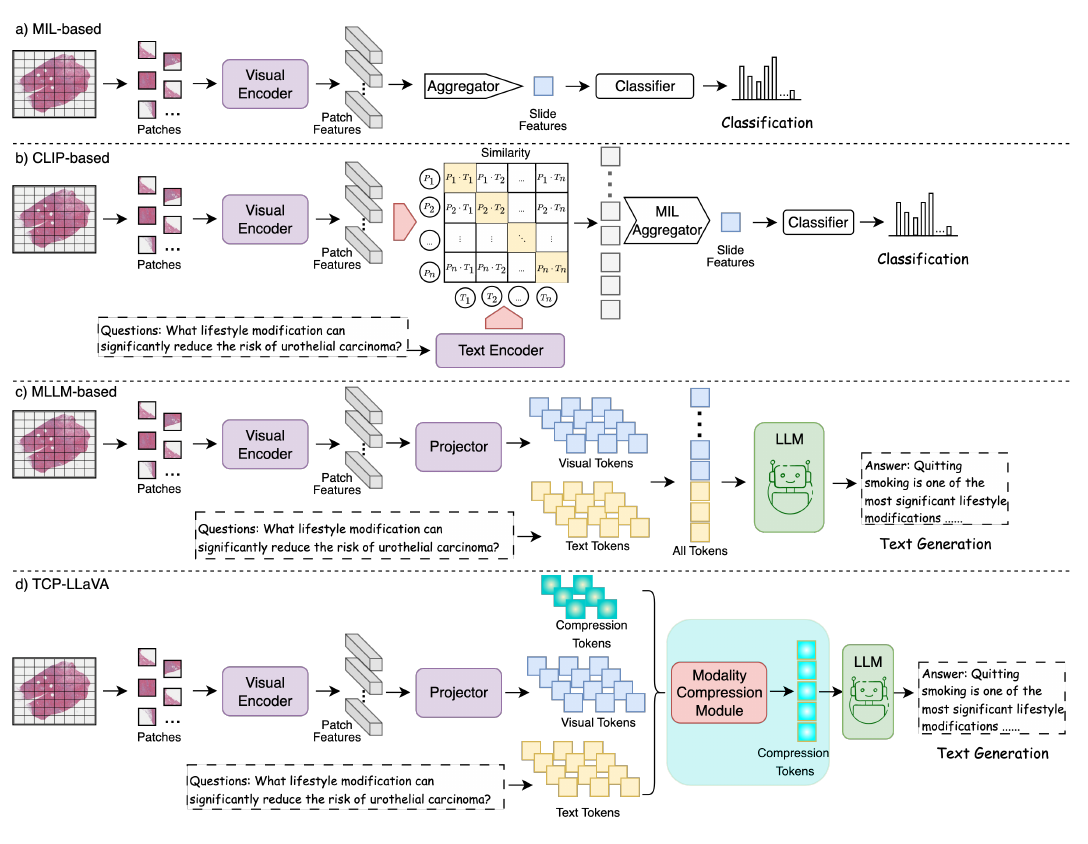}
  \vspace{-.15in}
  \caption{ Comparison of different WSI modeling paradigms for pathology tasks. 
(a) \textbf{MIL-based methods} aggregate patch-level features into a slide-level representation via aggregation modules, typically used for classification. 
(b) \textbf{CLIP-based methods} compute similarity scores between each patch and text prompt embeddings, followed by MIL aggregation for classification tasks. 
(c) \textbf{MLLM-based methods} directly feed all extracted visual tokens (often exceeding 10K) into a large language model (LLM) along with text tokens for end-to-end answer generation. 
(d) Our \textbf{TCP-LLaVA} introduces a modality compression module that distills thousands of patches and text tokens into a compact set of trainable compression tokens. This architecture significantly reduces computational load while maintaining high performance on gigapixel-scale VQA tasks.
 }
  \vspace{-.1in}  
  \label{fig:arch_diff}
\end{figure*}

With the advent of multimodal large language model (MLLM) like LLaVA~\citep{liu2023visual}, several recent studies have begun adapting these models for WSI-related text generation tasks. Unlike CLIP which employs dual encoders trained via contrastive objectives to align image and text embeddings, LLaVA extends CLIP’s frozen visual encoder by adding a projection module and integrating a full-scale large language model as the decoder. 
This design enables MLLM not only to align images with text but also generate descriptive textual outputs. Models such as SlideChat~\citep{chen2025slidechat} and CPath-Omni~\citep{guo2025focus} directly input all visual tokens, extracted from thousands of WSI patches, into the large language model, thereby enabling end-to-end VQA capabilities. However, this brute-force tokenization approach requires substantial computational resources due to the extremely long token sequences derived from gigapixel WSIs. Consequently, a clear research gap emerges regarding efficient methods for summarizing or compressing WSI inputs, particularly patch tokens, prior to LLM processing, aiming to preserve critical diagnostic information while significantly reducing computational overhead.

To address the computational challenges associated with gigapixel-scale WSI understanding in VQA tasks, we propose \textbf{Token Compression Pathology LLaVA (TCP-LLaVA)}, a novel multimodal large language model architecture designed for efficient processing of whole slide images. TCP-LLaVA effectively handles the extreme input sequence lengths by introducing a compact set of special compression tokens along with a lightweight yet powerful modality compression module. Our method consists of four primary stages: first, visual tokens are extracted at the patch level using a frozen visual encoder; second, text tokens are generated from the input prompt via a pretrained language tokenizer; third, visual and textual features are jointly fused and compressed into special compression tokens within the modality compression module; and finally, the large language model decodes these compressed tokens to generate free-form textual answers. This design significantly reduces computational demands while preserving the ability to reason effectively over large-scale WSIs.

The core innovation of TCP-LLaVA lies in the introduction of a fixed number of special, trainable compression tokens. These tokens, analogous to the [CLS] token in BERT, aggregate information from thousands of patch-derived visual tokens and the question tokens through a modality compression module that leverages multi-head attention mechanisms. Only the final hidden states of these compression tokens are passed forward to the large language model as concise yet informative representations of the WSI. By compressing extensive visual and textual inputs into a compact set of tokens, TCP-LLaVA effectively preserves critical diagnostic content while significantly reducing memory usage and computational demands, enabling practical feasibility for large-scale WSI VQA tasks. The contributions are summarized as follows:

\begin{itemize}

    \item We propose the first token-compression-based multimodal large language model, \textbf{TCP-LLaVA}, designed explicitly for whole slide image (WSI) visual question answering tasks.

    \item TCP-LLaVA introduces a compact set of special, trainable compression tokens along with an efficient modality compression module to effectively summarize both visual and textual inputs into fixed-length representations.

    \item TCP-LLaVA significantly reduces memory and computational costs associated with gigapixel-scale WSIs, achieving competitive VQA performance compared to state-of-the-art methods, but with substantially fewer tokens.

\end{itemize}








\section{Related Work}


\myfirstpara{Whole Slide Image Classification.}
Multiple Instance Learning (MIL) addresses the gigapixel scale of WSIs by dividing them into thousands of patches, treating the entire slide as a \textit{bag} and the patches as \textit{instances}, a dominant paradigm for WSI classification. Aggregation functions such as attention-based pooling~\citep{ilse2018attention} or CLAM~\citep{lu2021data} combine patch-level features into slide-level predictions, accounting for the fact that only a few patches may carry diagnostic signals. Recently, CLIP~\cite{radford2021learning} has been integrated into MIL to leverage image-text pretraining for stronger patch features~\cite{zhang2024biomedclip}. For example, ViLa-MIL~\cite{shi2024vila} uses a dual-scale design with text-guided prompts to refine patch representations. However, these methods produce a single label or low-dimensional embedding, which limits their suitability for complex VQA tasks.

\myfirstpara{Whole Slide Image Text Generation.}
Early efforts to move beyond classification focused on generating descriptive text or pathology reports from WSIs. These models typically follow an encoder-decoder architecture. For instance, HistoCap~\cite{sengupta2024automatic} utilizes a pre-trained Vision Transformer for histopathology (HIPT)~\cite{chen2022scaling} as the vision encoder to extract features from WSI patches and an LSTM-based decoder to generate captions. Similarly, Guevara et al.~\cite{guevara2023caption} also employ pre-trained transformers to generate captions for histopathology images. Another step towards more advanced generating descriptive text, WSI-VQA~\cite{chen2024wsi}, curates a bigger dataset and involves feeding all patch-derived visual tokens into an encoder-decoder transformer architecture. Howerver, since WSI-VQA~\cite{chen2024wsi} is only trained pathology dataset without leverage large scale pre-training, the generalization ability is limited. While these models represent a step towards interpreting WSIs, they are not designed for interactive, dialogue-based VQA and often struggle to generate detailed, clinically relevant reports that require reasoning across the entire slide.

\myfirstpara{Multimodal Large Language Model in WSI.}
MLLM~\cite{li2023llava,liu2023llava} architectures has opened the gate for generative AI for computational pathology. These models connect a pre-trained vision encoder to LLM ~\cite{liu2023llava}, enabling sophisticated VQA capabilities. However, due to the long-sequence modeling and computational challenge, recent work such as SlideChat~\cite{chen2025slidechat} deploys additional long-sequence module along side with the visual encoder to process all the visual patches at the slide-level. However, this brute-force approach brings computational overhead during training and inference. The long sequence could harm the performance of LLM. Other models like CPath-Omni~\cite{sun2025cpath} propose multi-scale feature extraction but still generate a vast number of tokens that are passed to the LLM. PathGen-1.6M~\cite{sun2024pathgen} focuses on generating a large-scale dataset of patch-caption pairs to pre-train a MLLM-style model, but it primarily operates at the patch level, missing global WSI context. The core inefficiency of these approaches for WSIs lies in the lack of an intelligent mechanism to summarize or compress the massive number of visual tokens before LLM processing, a gap that our proposed token-compression strategy directly addresses.







\section{Token Compression Pathology LLaVA}


\begin{figure*}[!t]
  \centering
 \vspace{-.1in}
  \includegraphics[width=\textwidth]{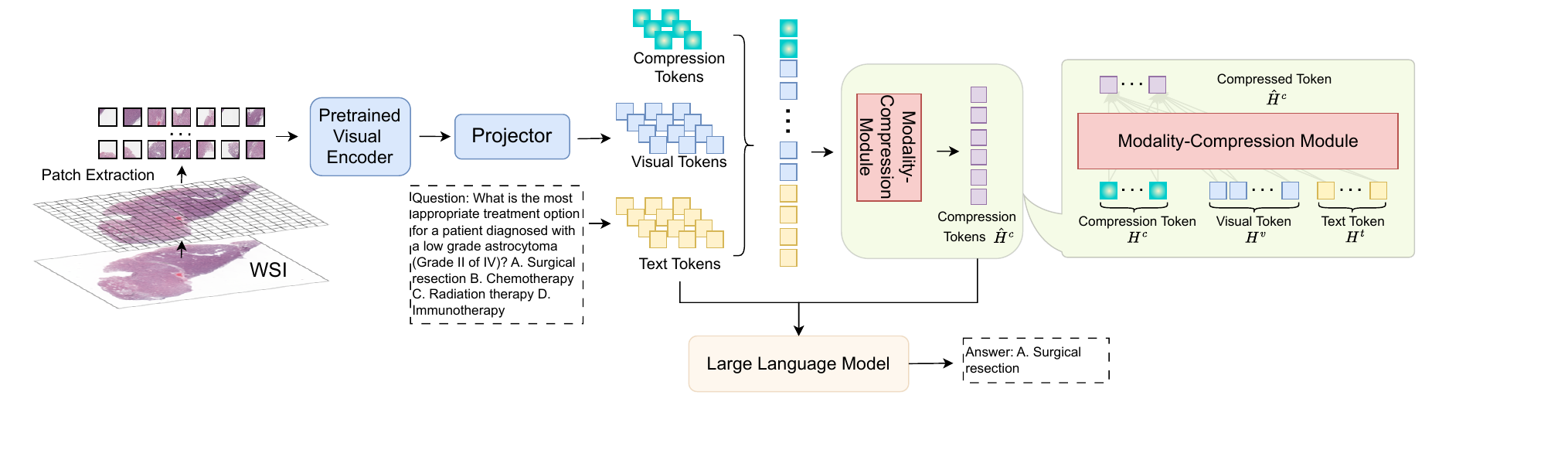}
  \vspace{-.3in}
  \caption{Overall architecture of \textbf{TCP-LLaVA} for visual question answering on whole slide images (WSIs). Given a high-resolution WSI, non-overlapping patches are extracted and passed through a pretrained visual encoder, followed by a projector that aligns visual features to the LLM embedding space, producing visual tokens. Meanwhile, the question and answer choices are tokenized into text tokens. These, along with a set of special trainable compression tokens, are input to the Modality Compression Module, which performs cross-modal attention to distill a compact representation. Only the updated compressed tokens are forwarded to the large language model (LLM) for answer generation. This design enables efficient and scalable reasoning on gigapixel WSIs while significantly reducing input sequence length and computational cost.}
  \vspace{-.1in}  
  \label{fig:framework}
\end{figure*}

In this section, we present \textbf{Token Compression Pathology LLaVA (TCP-LLaVA)}, a novel multimodal large language model (MLLM) architecture specifically designed for visual question answering (VQA) on gigapixel-scale whole slide images (WSIs). As illustrated in Fig.~\ref{fig:framework}, TCP-LLaVA is built to address the substantial sequence length challenges posed by WSIs. To this end, it incorporates a lightweight modality compression mechanism that enables efficient representation of dense visual content, significantly reducing input length while preserving diagnostic relevance.

Traditional MLLM-based approaches, such as SlideChat~\cite{chen2025slidechat}, rely on directly passing all patch-level visual tokens into the language model. Given that WSIs typically produce over 10,000 visual tokens per slide, this direct-forwarding strategy leads to prohibitively high memory usage and computational overhead, limiting their practical deployment. In contrast, TCP-LLaVA implements an early-stage cross-modal token compression step, condensing both visual and textual information into a fixed-length representation. This strategy achieves over $10\times$ input reduction and enables practical inference on high-resolution WSIs using standard hardware configurations.

Specifically, TCP-LLaVA comprises four sequential stages: (1) visual token extraction at the patch level using a frozen pretrained encoder; (2) text encoding via a pretrained LLM tokenizer; (3) fusion and compression of visual and textual features into a set of special trainable compression tokens within a dedicated modality compression module; and (4) generation of free-form textual answers by the large language model, solely based on these compressed token representations. This design effectively balances computational efficiency and model expressiveness, ensuring scalability to high-resolution WSIs. The subsequent subsections detail each architectural component, highlighting how they collectively support efficient and effective multimodal learning for pathology applications.

\subsection{Patch Extraction and Visual Token Encoding}

Following standard pathology preprocessing procedures~\cite{lu2021data}, each WSI is divided into non-overlapping patches of size $224 \times 224$ pixels. This tiling operation typically results in thousands to tens of thousands of patches per slide. Each patch is passed through a pretrained visual encoder (e.g., CONCH~\cite{lu2024visual}) to produce dense feature embeddings. These features are then projected via a projector into the same latent space as the language model, yielding a sequence of visual tokens:
\vspace{-.05in}
\[
H^v \in \mathbb{R}^{l_{\text{WSI}} \times d_h},
\]

\vspace{-.05in}
 
where $l_{\text{WSI}}$ denotes the number of extracted patches and $d_h$ is the shared hidden dimension. In practice, $l_{\text{WSI}}$ can exceed 10,000 tokens per slide.

\subsection{Text Token Encoding}

Given a visual question answering (VQA) prompt associated with the WSI, such as a question and its four choices, the text is tokenized using the language model's tokenizer and mapped into embeddings via the LLM’s input layer. This results in a sequence of text tokens:
\vspace{-.05in}
\[
H^t \in \mathbb{R}^{l_t \times d_h},
\]

\vspace{-.05in}

where $l_t$ is the number of tokens in the input question.

\subsection{Modality Compression Module}
\label{sec:modalitycompression}

\myparagraph{Motivation and Overview.}
To enable efficient reasoning over long visual and textual inputs from gigapixel WSIs, we introduce a \textit{Modality Compression Module}. This component functions as a cross-modal fusion mechanism that compresses thousands of visual and text tokens into a compact, fixed-length representation. Specifically, we prepend a set of special trainable compression tokens to the combined sequence of visual and text tokens. These compression tokens, together with the fusion mechanism in the modality compression module, play a role similar to the [CLS] token in BERT models—learning to aggregate and summarize information across the entire input. After fusion, only the final hidden states of the compression tokens are forwarded to the language model decoder. This selective processing significantly reduces computational cost while preserving essential, task-relevant semantic content.

\myparagraph{Trainable Compression Tokens.}
We define a set of special trainable \textit{compression tokens}:
\vspace{-.05in}
\[
H^c \in \mathbb{R}^{l_c \times d_h},
\]

\vspace{-.05in}

where $l_c$ is a small, pre-defined number (e.g., 100), and $d_h$ is the hidden dimension shared across modalities. These tokens are randomly initialized and optimized during training. Conceptually, they function analogously to the [CLS] token in BERT~\cite{devlin2019bert}, each compression token learns to attend over the full sequence of visual and textual inputs, serving as a dynamic summary vector that aggregates salient cross-modal features into a concise format.

\myparagraph{Cross-Modality Fusion.}
Let $H^v \in \mathbb{R}^{l_v \times d_h}$ denote the projected visual tokens, $H^t \in \mathbb{R}^{l_t \times d_h}$ the embedded text tokens, and $H^c \in \mathbb{R}^{l_c \times d_h}$ the initialized trainable compression tokens. We concatenate these three components to form a joint sequence: $(H^c, H^v, H^t)$,
 and feed them into the modality compression module. The updated compression tokens are computed as:

\vspace{-.05in}
\[
\hat{H}^c = \text{ModalityCompression}( \text{Concat}(H^c, H^v, H^t) )[:l_c],
\]

\vspace{-.05in}

where $\hat{H}^c \in \mathbb{R}^{l_c \times d_h}$ represents the fused compression tokens. Crucially, only the resulting $\hat{H}^c$ is forwarded to the language model decoder. This selective forwarding ensures that the LLM operates on a distilled representation of the WSI and the associated question, substantially improving memory and inference efficiency. The number of compression tokens $\hat{H}^c$ can be as less as 100 tokens. Each compression token employs multi-head attention to query the full visual and textual context, learning to extract semantically rich and task-relevant information.




\myparagraph{Layer Initialization for Improved Alignment.} 
To further enhance fusion quality, we initialize the modality compression module using the early layers of the target LLM. Prior work~\cite{fastv2024liang,huang2025dynamicllava} suggests that lower layers of LLMs capture fundamental syntactic structure and alignment signals. By reusing these layers, our model gains an inductive bias that improves semantic alignment between modalities and accelerates training convergence. This initialization strategy helps guide the compression tokens to focus on content-rich regions across both image and text inputs during fusion.

\subsection{LLM Decoding for VQA}
The final stage of TCP-LLaVA involves answer generation using a large language model (LLM) decoder. The decoder receives the compressed token representations $\hat{H}^c$, produced by the modality compression module, along with the original question prompt. The prompt includes both the question and its corresponding multiple-choice options, as illustrated in Fig.~\ref{fig:sankey_diagram}. We adopt a \textbf{free-form generation} setting, where the LLM generates natural language responses (e.g., ``A. Surgical resection'') rather than selecting from predefined answer choices. This generative formulation offers greater flexibility and expressiveness, making it well-suited for modeling complex diagnostic reasoning and descriptive interpretation tasks in computational pathology.

\subsection{Training}
\label{sec:training}

Each training sample consists of a whole slide image $I$, a natural language question $Q$, and a corresponding answer $A$. The objective is to generate accurate, free-form answers conditioned on both the image and the question.

To ensure stable training and efficient resource usage, we adopt a partial fine-tuning strategy. Specifically, the visual encoder and the LLM are kept frozen throughout training. Only the \textit{projection module} and the \textit{Modality Compression Module} are updated. This design not only reduces the number of trainable parameters but also accelerates convergence and further lowers computational costs.

\myparagraph{Training Objective.}
Let $A = (a_1, a_2, \ldots, a_T)$ denote the ground-truth answer tokens. The model is trained with a standard autoregressive language modeling objective. Given the compressed multimodal representation $\hat{H}^c$ and the question $Q$, we minimize the following negative log-likelihood loss:
\vspace{-.05in}

\[
\mathcal{L}_{\text{VQA}} = -\sum_{t=1}^{T} \log P(a_t \mid a_{<t}, \hat{H}^c, Q),
\]

\vspace{-.05in}

where $\hat{H}^c$ is the compressed output from the Modality Compression Module described in Sec.~\ref{sec:modalitycompression}, $P(a_t \mid \cdot)$ is the probability of the $t$-th token generated by the LLM, $I$ denote a whole slide image, and $Q$ denotes the question.

\myparagraph{Training Efficiency.}
TCP-LLaVA is designed to significantly reduce the memory and compute burden associated with processing high-resolution WSIs. In standard MLLM pipelines such as SlideChat~\citep{chen2025slidechat}, the full sequence of patch-level visual tokens-often exceeding 10,000—is forwarded into the LLM, leading to high memory consumption and slow training. In contrast, our model replaces these with a compact set of 100 compression tokens, reducing the LLM input length by over \textbf{99\%}.

This token reduction yields substantial improvements in throughput and time-to-convergence. As detailed in Sec.~\ref{sec:training_inference_efficiency}, TCP-LLaVA achieves higher TFLOPS and samples-per-second rates during training, and requires fewer GPU hours to reach comparable accuracy levels. These gains are particularly important when scaling to large-scale pathology datasets with gigapixel image inputs.

\section{Experiments}

\begin{figure*}[!ht]
  \centering
 \vspace{-.1in}
  \includegraphics[width=\textwidth]{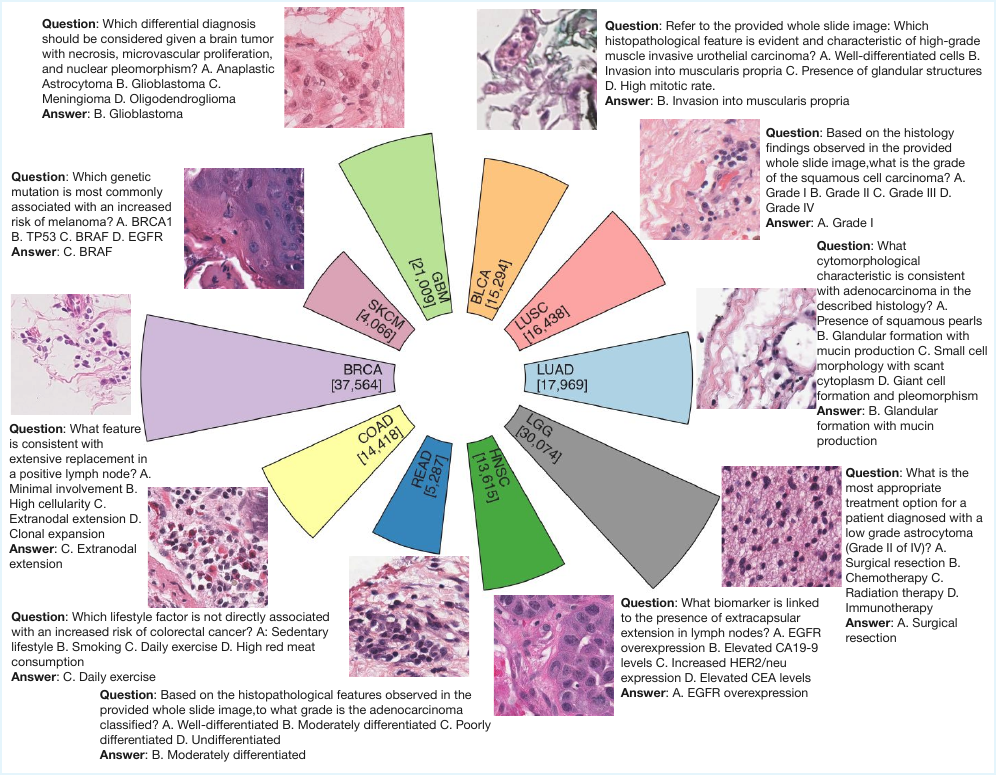}
  \vspace{-.15in}
  \caption{Visualization of the tumor-type distribution in our curated multi-tumor VQA benchmark dataset, constructed from TCGA~\citep{TCGA_GDC} and refined with annotations from SlideBench~\citep{chen2025slidechat}. Each colored segment represents one of ten tumor types, with the arc length proportional to the number of associated question-answer (QA) pairs. The dataset includes BRCA (37,564), LGG (30,074), COAD (14,481), HNSC (13,615), GBM (21,009), LUAD (17,969), LUSC (16,438), BLCA (15,294), SKCM (4,066), and READ (5,287). For each tumor type, we illustrate a representative whole slide image (WSI) and a corresponding clinical question-answer pair, highlighting the dataset’s diversity and the depth of pathology-informed reasoning. 
  }
  \vspace{-.1in}  
  \label{fig:sankey_diagram}
\end{figure*}

\subsection{Multi-Tumor TCGA QA Benchmark Dataset}

In this work, we construct a diverse VQA dataset based on the publicly available The Cancer Genome Atlas (TCGA) dataset~\citep{TCGA_GDC} by integrating and refining annotations from SlideBench~\citep{chen2025slidechat} and WSI-VQA~\citep{chen2024wsi}. 

Prior VQA datasets do not explicitly label the tumor types associated with each WSI. To facilitate tumor-specific analysis and evaluate model generalization across cancer types, we categorize the data into ten tumor subtypes: BLCA, BRCA, COAD, GBM, HNSC, LGG, LUAD, LUSC, READ, and SKCM. Each subtype includes a set of WSIs along with natural language question-answer pairs tailored to clinical and pathological reasoning. Summary statistics of the subtype distribution are shown in Fig.~\ref{fig:sankey_diagram}. Detailed tumer type specific statistics can be found in Appx.~\ref{app:experimental}.

Following the protocol in \citet{chen2025slidechat}, we split the WSI data within each tumor type into training, validation, and test sets using an 8:1:1 ratio. This ensures that no whole slide image (WSI) appears in more than one split, thereby avoiding data leakage. The splitting strategy preserves the distribution of each tumor subtype, enabling robust evaluation while also enhancing subtype granularity for more detailed analysis.

\subsection{Experimental Settings}

\myparagraph{Baselines.}
We compare TCP-LLaVA with several representative MLLM-based baselines across four technical categories. (1) \textit{General-purpose MLLMs}: LLaVA-v1.6-Vicuna-7B~\cite{liu2023llava}, a widely used open-domain vision-language model. (2) \textit{Token pruning methods}: DivPrune~\cite{Alvar_2025_CVPR}, which reduces the token length via diversity-aware pruning. (3) \textit{Patch-level MLLMs}: LLaVA-Med~\citep{li2023llava} and Quilt-LLaVA~\citep{seyfioglu2024quilt}, which use isolated patch inputs for question answering. (4) \textit{WSI-level MLLMs}: SlideChat~\cite{chen2025slidechat}, which directly feeds thousands of patch tokens into the LLM. For the general-purpose and patch-level models, we follow SlideChat’s setup and randomly sample 30 patches per slide as visual inputs. For WSI-level models and pruning-based methods, we cap the visual token sequence length at 10,000 to prevent out-of-memory errors during training and inference.

\myparagraph{Training Settings.}
For TCP-LLaVA, the visual encoder is initialized from CONCH~\cite{lu2024visual}, while the projector and language model\footnote{The LLM is instantiated from the Qwen2.5-7B-Instruct architecture~\cite{yang2024qwen2technicalreport}.} are initialized from SlideChat~\cite{chen2025slidechat}, which employs a two-stage training strategy to enhance WSI understanding. During training, we freeze both the visual encoder and the LLM, and update only the modality compression module and the projection layer. \textit{Notably, under this partial fine-tuning setup, training TCP-LLaVA on 10,000 samples takes approximately 0.67 hours. In comparison, SlideChat requires about 1 hour to train under the same conditions while updating only the projector.}

\myparagraph{Evaluation Metrics.}
We evaluate our model using \textbf{accuracy}, the standard metric for multiple-choice VQA tasks. Each question in the dataset is associated with four answer options (A–D), with only one correct answer. The model is prompted to generate a free-form response (e.g., ``A. Surgical resection''). During evaluation, we extract the model's predicted choice letter (e.g., ``A'') using a rule-based parser and compare it against the ground-truth label. A prediction is considered correct if the extracted choice matches the annotated answer key. Final accuracy is computed as the proportion of correctly answered questions across the evaluation set.

\begin{table*}[!t]
  \caption{Performance comparison across ten tumor types on the TCGA benchmark. We report VQA accuracy (\%) for various LLaVA-based methods and our proposed TCP-LLaVA. TCP-LLaVA achieves the highest average accuracy (78.57\%) across all tumor types, demonstrating the effectiveness of token compression for enhancing both performance and scalability in gigapixel-scale WSI VQA.}
  \label{tab:main_tcga_table}
  \centering
  \vspace{-.1in}
    \resizebox{\textwidth}{!}{ 
\begin{tabular}{c|cccccccccc|c}
\toprule
\textbf{Methods/Cancer Type}  & \textbf{BLCA} & \textbf{BRCA} & \textbf{COAD} & \textbf{GBM} & \textbf{HNSC} & \textbf{LGG} & \textbf{LUAD} & \textbf{LUSC} & \textbf{READ} & \textbf{SKCM} & \textbf{AVG} \\ \midrule
\textbf{LLaVa-v1.6-Vicuna-7b} & 32.38         & 28.41         & 36.94         & 25.00        & 35.35         & 36.50        & 43.46         & 34.46         & 47.30         & 26.47         & 34.63        \\
\textbf{LLaVA-Med}            & 46.84         & 37.58         & 52.23         & 40.00        & 47.47         & 40.88        & 29.84         & 51.41         & 64.86         & 32.35         & 44.35        \\
\textbf{Quilt-LLaVA}          & 37.97         & 28.18         & 31.21         & 45.00        & 41.41         & 37.23        & 30.37         & 29.94         & 52.70         & 55.88         & 38.99        \\ \midrule
\textbf{SlideChat}            & 81.65         & 69.57         & 74.52         & 75.00        & 78.79         & 79.34        & 74.87         & 74.01         & 85.14         & 79.41         & 77.23        \\
\textbf{SlideChat + DivPrune} & 82.91         & 69.57         & 73.89         & 75.00        & 78.79         & 79.34        & 74.87         & 72.88         & 85.14         & 79.41         & 77.18        \\ \midrule
\textbf{TCP-LLaVA}            & 83.54         & 67.56         & 76.43         & 80.00        & 78.79         & 79.34        & 76.44         & 73.45         & 87.84         & 82.35         & 78.57        \\ \bottomrule
\end{tabular}
}
\vspace{-.05in}  
\end{table*}

\myparagraph{Implementation Details.}
We implement TCP-LLaVA using PyTorch. We train the TCP-LLaVA with NVIDIA A6000 (48G GPU memory) GPUs. We use AdamW optimizer with a linear warmup and cosine decay schedule with initial learning rate $1.5e-5$. The model is fine-tuned for 2 epochs with a batch size of 1. We also use $8$ gradient accumulation steps and mixed precision (FP16) to reduce memory consumption.


\subsection{Results}

Tab.~\ref{tab:main_tcga_table}\footnote{We also implemented the WSI-VQA baseline~\citep{chen2024wsi}, but it achieved 0\% accuracy on our benchmark. This poor performance is primarily attributed to two factors: (1) the model employs a hand-crafted tokenizer specifically tailored to a narrow pathology vocabulary, which generalizes poorly to our multi-tumor TCGA VQA dataset—frequently mapping key medical terms to the \texttt{[UNK]} token; and (2) the original WSI-VQA was developed primarily for the BRCA tumor type and lacks mechanisms for handling diverse tumor subtypes. These limitations critically impair the model’s ability to comprehend and respond accurately to questions in our broader benchmark setting.} reports VQA accuracy across ten TCGA tumor types for a suite of MLLM methods. Off‑the‑shelf baselines (LLaVa‑v1.6‑Vicuna‑7b, LLaVA‑Med, Quilt‑LLaVA) perform poorly (34.63\%, 44.35\%, and 38.99\% average accuracy), underscoring that generic LLMs struggle on WSI. By contrast, SlideChat, our initial pipeline, yields a large jump to 77.23\% average accuracy, showing that explicit image reasoning and domain‑specific prompting are key. Incorporating the token pruning method, DivPrune, into SlideChat maintains this level (77.18\%), confirming that aggressive token reduction slightly sacrifices accuracy.

Our TCP‑LLaVA achieves the best performance at 78.57\% average accuracy. TCP‑LLaVA consistently outperforms SlideChat across most of tumor types (e.g., +1.89 pp\footnote{``pp'' stands for percentage points, which denotes the absolute difference between two percentages. For example, on BLCA, SlideChat scores 81.65\% accuracy and TCP‑LLaVA scores 83.54\%, an increase of 1.89 percentage points (pp), not a 1.89\% relative gain over the original value.} on BLCA, +1.91 pp on COAD, +1.57 pp on LUAD, +2.70 pp on READ, and +2.94 pp on SKCM). It achieves its largest improvement on challenging tumor types, i.e., GBM, with a +5.00 pp gain.

These results demonstrate that (1) modality compression can preserve or even boost the MLLM performance on WSI VQA tasks, and (2) our proposed TCP‑LLaVA framework establishes a new state of the art for histopathology question answering on the TCGA benchmark.


\begin{figure}[h]
  \centering
  \includegraphics[width=\columnwidth]{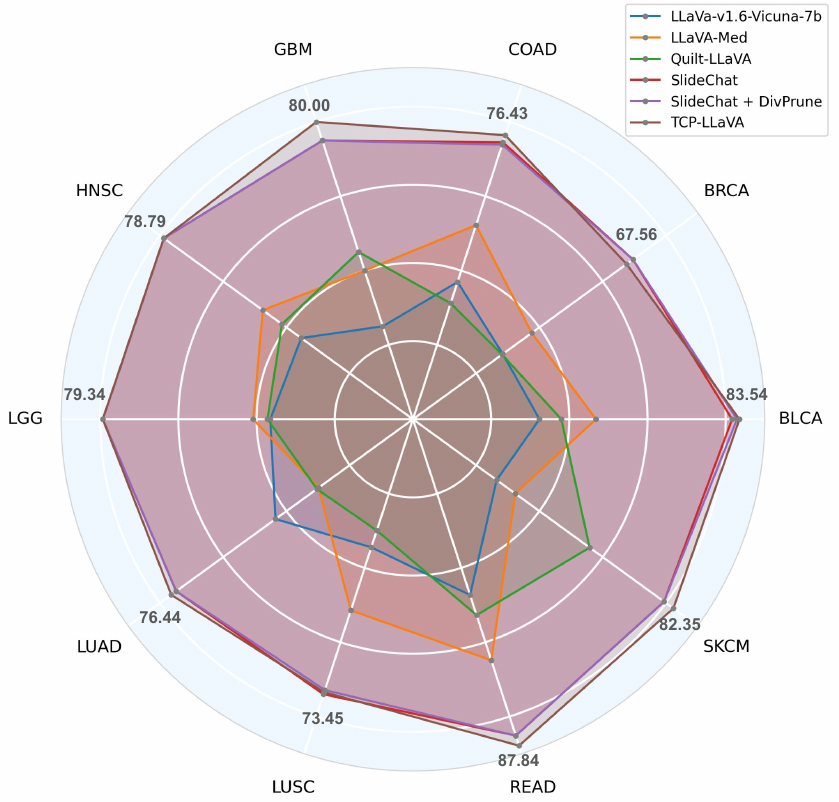}
  \vspace{-.1in}
  \caption{Radar chart of the performance comparison across ten tumor types on the TCGA benchmark. The accuracy of TCP-LLaVA is illustrated for each tumor type. }
  \vspace{-.1in}  
  \label{fig:radar_chart}
\end{figure}

\subsection{Training and Inference Efficiency}
\label{sec:training_inference_efficiency}

To assess the computational efficiency of our TCP-LLaVA, we evaluate both training and inference using two metrics: \textbf{TFLOPS} and \textbf{Throughput (samples/sec)}. TFLOPS captures the model's raw computational utilization, while throughput reflects its practical processing speed in real-world settings. Given variability in input lengths across WSIs and questions, we report the average values of both metrics. Formal definitions and further details are provided in Appx.~\ref{app:efficiency_metrics}.

In our experiments, TCP-LLaVA employs 100 compression tokens as the intermediate representation, while the SlideChat baseline~\citep{chen2025slidechat} utilizes the full set of visual tokens, following its original design. The total number of visual tokens in SlideChat can exceed 10,000 per WSI, leading to substantially longer sequences being fed into the language model. By contrast, TCP-LLaVA reduces the input token length by over \textbf{99\%}, leading to improved GPU utilization and significantly lower training and inference times.


As reported in Tab.~\ref{tab:avg_tflops}, TCP-LLaVA achieves an average of \textbf{10.87 TFLOPS}, compared to \textbf{2.35 TFLOPS} for SlideChat. This corresponds to \textbf{over ~$4\times$ improvement in computational throughput}, indicating more efficient use of GPU resources. Similarly, TCP-LLaVA achieves much higher sample-level throughput during both training and inference phases, demonstrating the benefits of token compression for large-scale WSI-VQA tasks.

These results validate that the proposed compression-based architecture substantially improves computational efficiency, reducing resource consumption and training time without compromising model performance.

\begin{table}[h]
\centering
\caption{Comparison of average training throughput (TFLOPS) and throughput (samples/sec) on a single NVIDIA A6000 GPU with only batch size 1 and no gradient accumulation steps.}
\vspace{-.1in} 
\label{tab:avg_tflops}
\resizebox{\columnwidth}{!}{
\begin{tabular}{c|cc|c}
\toprule
\multirow{2}{*}{\textbf{Model}} & \multicolumn{2}{c|}{\textbf{Training}}                  & \textbf{Inference}                \\ \cmidrule(l){2-4} 
                                & \textbf{TFLOPS\ua} & \textbf{Throughput (samples/sec)\ua} & \textbf{Throughput (samples/sec)\ua} \\ \midrule
\textbf{SlideChat}              &         2.35        & 0.42                          & 0.58                           \\
\textbf{TCP-LLaVA}              & 10.87              & 179.46                           & 3.33                         \\ \bottomrule
\end{tabular}
}
\vspace{-.05in} 
\end{table}

\subsection{Ablation Study}

    





\myparagraph{Ablation of Token Compression vs. MIL Aggregation.}
To further evaluate the efficiency of our token compression approach, we conduct an ablation study comparing it with a widely used multiple instance learning (MIL) strategy. MIL-based methods are commonly adopted in computational pathology to aggregate patch-level features into slide-level representations. In this experiment, we integrate the ACMIL module~\citep{zhang2024attention}, a lightweight and effective MIL framework, into the SlideChat architecture. Specifically, we insert ACMIL between the visual projector and the LLM, allowing it to pool patch features before decoding.

As shown in Tab.~\ref{tab:ablation_efficiency}, TCP-LLaVA achieves better performance than ACMIL+SlideChat, particularly on the LUAD subtype. This demonstrates that our trainable token compression method is competitive in accuracy. Unlike MIL, which relies on deterministic pooling mechanisms, our approach enables task-driven feature aggregation through end-to-end learning, preserving richer cross-modal interactions.



\begin{table}[!t]
  
  \caption{Comparison between MIL-based aggregation (ACMIL+SlideChat) and our token compression approach (TCP-LLaVA) on BRCA and LUAD subtypes. TCP-LLaVA achieves better or comparable accuracy with improved scalability.}
  \label{tab:ablation_efficiency}
  \centering
  \small
  \vspace{-.1in}
\begin{tabular}{c|cc}
\toprule
                         & \textbf{BRCA} & \textbf{LUAD} \\ \midrule
\textbf{ACMIL+SlideChat} & 67.56         & 74.35         \\
\textbf{TCP-LLaVA}       & 67.56         & 75.39         \\ \bottomrule
\end{tabular}

\vspace{-.05in}  
\end{table}

\myparagraph{Ablation of Compression Token Numbers.}
Tab.~\ref{fig:ablation_compression_token_number} illustrates the impact of varying the number of compression tokens (100–4,000) on VQA accuracy for two representative tumor types. TCP‑LLaVA is highly robust across token budgets: BRCA accuracy remains nearly constant (67.1–67.8\%), while LUAD improves only modestly from 74.35\% at 100 tokens to 75.92\% at 4,000 tokens. These results indicate that, although increasing the token budget can yield marginal gains on certain tumors, a moderate setting (e.g., 100 tokens) already achieves an excellent trade‑off between accuracy and computational cost.

\begin{table}[]
\centering
\caption{Ablation study on the number of compression tokens for TCP-LLaVA. The model maintains stable performance across a wide range of token counts, demonstrating robustness and efficiency.}
\label{fig:ablation_compression_token_number}
\small
\vspace{-.1in} 
\begin{tabular}{c|cc}
\toprule
\textbf{Compression Token Numbers} & \multicolumn{1}{l}{\textbf{BRCA}} & \multicolumn{1}{l}{\textbf{LUAD}} \\ \midrule
\textbf{100}                       & 67.79                             & 74.35                             \\
\textbf{500}                       & 67.79                             & 75.39                             \\
\textbf{1000}                      & 67.11                             & 74.87                             \\
\textbf{2000}                      & 67.56                             & 75.39                             \\
\textbf{4000}                      & 67.79                             & 75.92                             \\ \bottomrule
\end{tabular}
\vspace{-.05in} 
\end{table}




\section{Conclusion}

In this paper, we introduced \textbf{Token Compression Pathology LLaVA (TCP-LLaVA)}, a novel multimodal large language model framework designed for visual question answering (VQA) on gigapixel-scale whole slide images (WSIs). To tackle the challenges posed by the extreme sequence lengths in WSI-level modeling, we proposed a modality compression module that distills visual and textual inputs into a compact set of informative tokens. This design significantly reduces memory and computational overhead while maintaining strong diagnostic performance.

Through extensive experiments on a curated multi-tumor TCGA QA benchmark, TCP-LLaVA achieves state-of-the-art performance across ten tumor types and demonstrates superior training and inference efficiency compared to existing MLLM-based baselines. Our architecture supports end-to-end training and scales effectively to full-resolution pathology slides on standard hardware.

\myparagraph{Limitations and Future Work.}
We believe TCP-LLaVA offers a promising direction for scaling multimodal foundation models to real-world clinical scenarios. While TCP-LLaVA provides an effective framework for WSI-based VQA, it currently focuses on question answering. In future work, we aim to extend the model toward more open-ended and generative tasks, such as pathology report generation. Or consider reinforcement learning frameworks ~\cite{cheng2023look, xu2022policy, jiang2024hummer, xu2023offline} to improve performance. A key challenge will be the creation of high-quality, fine-grained supervision reports to support this next stage of development.

\bibliographystyle{ieeenat_fullname}
\bibliography{main}


\clearpage
\newpage

\appendix

\section{Experimental Settings}
\label{app:experimental}

\myparagraph{TCGA Tumor Type statistics.} To provide a detailed understanding of the dataset composition, we summarize the number of WSIs and corresponding question-answer (QA) pairs for each tumor type as follows: BLCA contains \textbf{417} WSIs and \textbf{15294} QA pairs; BRCA contains \textbf{1050} WSIs and \textbf{37564} QA pairs; COAD contains \textbf{396} WSIs and \textbf{14418} QA pairs; GBM contains \textbf{504} WSIs and \textbf{21009} QA pairs; HNSC contains \textbf{423} WSIs and \textbf{13615} QA pairs; LGG contains \textbf{708} WSIs and \textbf{27749} QA pairs; LUAD contains \textbf{491} WSIs and \textbf{17917} QA pairs; LUSC contains \textbf{458} WSIs and \textbf{16438} QA pairs; READ contains \textbf{155} WSIs and \textbf{5287} QA pairs; and SKCM contains \textbf{105} WSIs and \textbf{4066} QA pairs.

\section{Training and Inference Efficiency Metrics}
\label{app:efficiency_metrics}

In Sec.~\ref{sec:training_inference_efficiency}, we introduce the computational efficiency of our TCP-LLaVA, and we evaluate with two metrics: \textbf{TFLOPS} and \textbf{Throughput (sample/sec)}. 
\textbf{TFLOPS}~\citep{towardsai.tflops}, or Tera–Floating Point Operations Per Second, quantifies the rate at which a GPU executes floating-point operations, providing an estimate of the effective computational performance of the model. \textbf{Throughput}~\citep{gyawali2023comparative}, defined as the number of WSI and QA samples processed per second, offers a direct measure of system efficiency in real-world training and inference scenarios.

Due to variability in input lengths—caused by differences in the number of visual tokens per slide, question token lengths, and output sequence lengths—we report the \emph{average} TFLOPS and throughput across the dataset. This allows for a fair and consistent comparison between models.

\section{Interpretability of the Modality Compression Module}

\begin{figure}
    \centering
    \includegraphics[width=1.1\columnwidth]{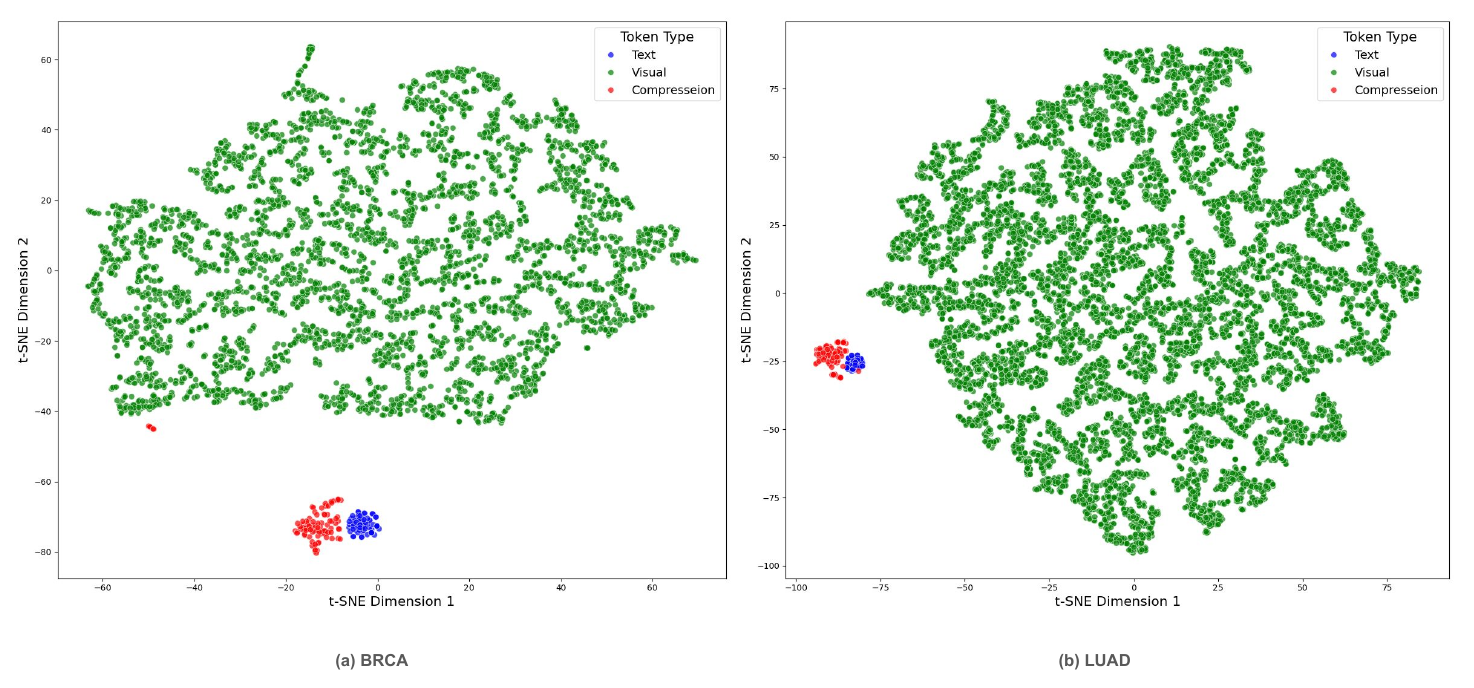}
    \caption{t-SNE visualization of three type tokens: uncompressed visual tokens, text tokens and compressed tokens. In (a), we visualize the BRCA subtype and in (b), we visualize the LUAD subtype.  }
    \label{fig:token_tsne}
\end{figure}

In order to demonstrate the information that compressed in the compression tokens, we plot the t-SNE of three types of tokens in Fig.~\ref{fig:token_tsne}. 

The first impression of Fig.~\ref{fig:token_tsne} is that green cloud of visual information has been successfully distilled into the tighter red cluster. The model has effectively summarized thousands of diverse visual features into a concise representation.

The compression tokens (red) are located right next to the blue text cluster and very far away from the original visual tokens in Fig.~\ref{fig:token_tsne}. This strongly suggests that the modality compression process is heavily guided by the text prompt. The model isn't just making a generic summary of the image; it's creating a summary specifically tailored to answer the question it was given.

Therefore, the design of the modality compression module does achieve the goal to effectively reduce the length of the visual tokens as well as integrate the text information into the compression tokens.

\end{document}